\documentclass[runningheads]{llncs}
\usepackage{graphicx}
\begin{document}
\title{On-Device Spatial Attention based Sequence Learning Approach for Scene Text Script Identification}
%
\titlerunning{On-Device Script Identification}
%
\author{Rutika Moharir\inst{1}
\and
Arun D Prabhu\inst{1} 
\and
Sukumar Moharana\inst{1}
\and
Gopi Ramena\inst{1}
\and
Rachit S Munjal\inst{1}}
\authorrunning{R. Moharir et al.}
%
\institute{On-Device AI, Samsung R\&D Institute, Bangalore, India \\
\email{ \{r.moharir, arun.prabhu, msukumar, gopi.ramena, rachit.m\}@samsung.com}}

\maketitle              
\begin{abstract}
Automatic identification of script is an essential component of a multilingual OCR engine. In this paper, we present an efficient, lightweight, real-time and on-device spatial attention based CNN-LSTM network for scene text script identification, feasible for deployment on resource constrained mobile devices. Our network consists of a CNN, equipped with a spatial attention module which helps reduce the spatial distortions present in natural images. This allows the feature extractor to generate rich image representations while ignoring the deformities and thereby, enhancing the performance of this fine grained classification task. The network also employs residue convolutional blocks to build a deep network to focus on the discriminative features of a script. The CNN learns the text feature representation by identifying each character as belonging to a particular script and the long term spatial dependencies within the text are captured using the sequence learning capabilities of the LSTM layers. Combining the spatial attention mechanism with the residue convolutional blocks, we are able to enhance the performance of the baseline CNN to build an end-to-end trainable network for script identification. The experimental results on several standard benchmarks demonstrate the effectiveness of our method. The network achieves competitive accuracy with state-of-the-art methods and is superior in terms of network size, with a total of just 1.1 million parameters and inference time of 2.7 milliseconds.

\end{abstract}
\section{Introduction}
\label{sec:intro}
Most multilingual OCR systems require knowledge of the underlying script for each text instance present in any natural image. Hence, script identification acts as an essential step in increasing the overall efficiency of any text recognition pipeline. Many languages
share the same script family, which makes script identification a fine-grained classification
task.

Earlier works in script identification have been proposed for document \cite{joshi2007generalised}, \cite{busch2005texture}, handwritten text \cite{hangarge2010offline}, \cite{hochberg1999script} and video-overlaid text \cite{zhao2012new}, \cite{gllavata2005script} where the texts are fairly simple with non-complex backgrounds and have been able to achieve great performance. However, scene text images pose additional challenges such as varied appearances and distortions, various text styles and degradation like background noise and light conditions, etc. Our work on scene text script identification focuses on two challenges – first, handling spatial distortions in scene text and exploiting sequential dependencies within the text which can help focus on discriminative features for efficient classification of scripts with relatively subtle differences, for example  Chinese and Japanese which share a common subset of characters. Second, be suitable, have low latency and memory footprint, for on-device deployment.

\begin{figure}[b]
\begin{center}
  \includegraphics[width=\textwidth,height=80 cm,keepaspectratio]{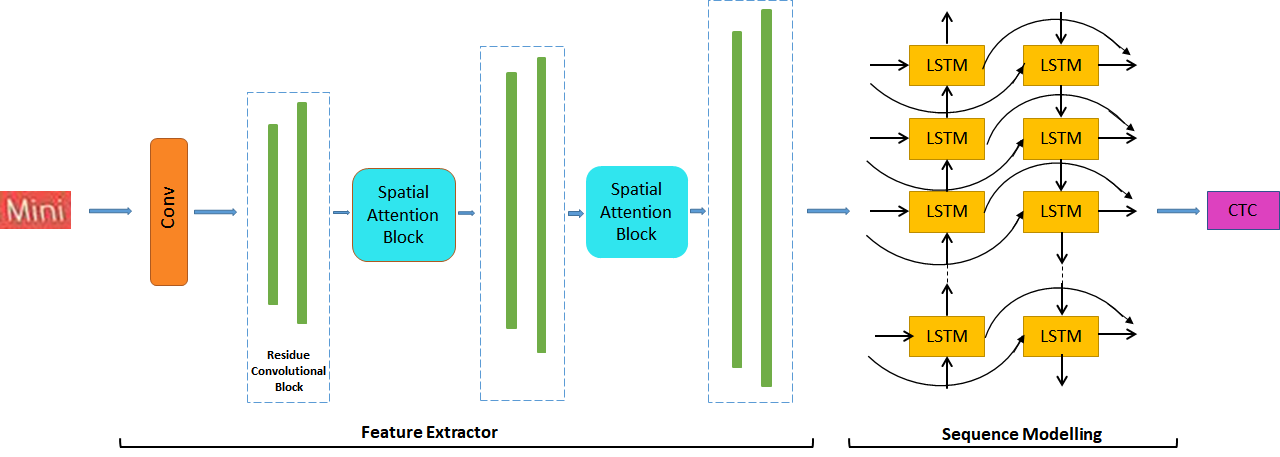}
\end{center}
   \caption{The architecture of the proposed script identification model.}
\label{fig:pipeline}
\end{figure}

\begin{figure}[t]
\begin{center}
  \includegraphics[width=\textwidth,height=\textheight, keepaspectratio]{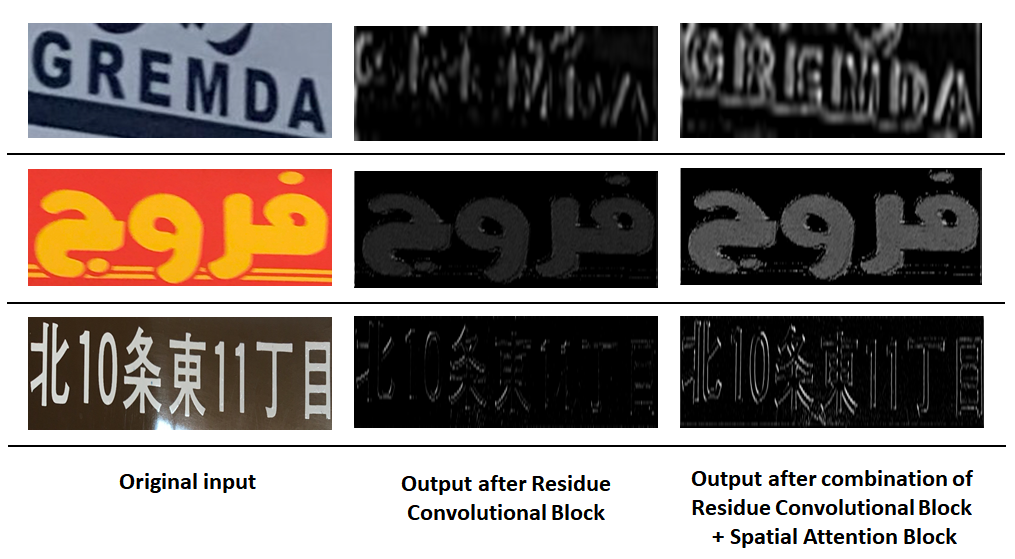}
\end{center}
 \caption{Illustration of the effect of the Spatial Attention Block during feature extraction. The 2nd column represents feature maps generated in the absence of the attention blocks while the 3rd column shows enhanced foreground-background separation with the removal of distortions on the application of the attention blocks.}
\label{fig:Blocks}
\end{figure}

While recent works have been trying to handle script identification in complex backgrounds \cite{bhunia2019script}, \cite{buvsta2018e2e}, \cite{zdenek2017bag}, \cite{fujii2017sequence}, \cite{cheng2019patch}, they have their own challenges. \cite{cheng2019patch} tries to enhance discriminative features using intermediate supervision on patch-level predictions. While, \cite{bhunia2019script} proposed an attention-based Convolutional-LSTM network, analyzing features globally and locally. \cite{buvsta2018e2e} uses an end-to-end trainable single fully convolutional network (FCN) while \cite{gomez2016fine} adopted a patch-based method containing an ensemble of identical nets to learn discriminative strokepart representations. In \cite{zdenek2017bag} they used BLCT to enhance the discriminative features and imposed inverse document frequency (idf) to get codewords occurrences. \cite{fujii2017sequence} use Encoder and Summarizer to get local features and fuse them to a single summary by attention mechanism to reflect the importance of different patches. However, few works have been reported to address an important factor that often leads to script identification failures – the degradation of scene text such as blur, tilt, low contrast to the background, etc. which are not rare in natural images. Instead of using computationally expensive attention-based patch weights \cite{bhunia2019script} or optimizing patch level predictions \cite{cheng2019patch}, we tackle this problem using a spatial attention mechanism that is capable of directly learning the region of interest for each character in the text for a superior foreground-background separation. Figure  \ref{fig:Blocks} shows the effect of using our spatial attention mechanism. The distorted text images are transformed and made suitable for script identification. 

As for the problem of failing to capture discriminative features, \cite{nicolaou2016visual} presented a method based on hand-crafted features, a LBP variant, and a deep multi-Layer perceptron. Pre-defined image classification algorithms, such as \cite{mei2016scene} who proposed a combination of CNN and LSTM model, the robustly tested CNN \cite{shi2016script} or the Single-Layer Networks (SLN) \cite{sharma2015icdar2015} normally consider holistic representation of images. Hence they perform poorly in distinguishing similarly structured script categories as they fail to take into account the discriminative features of the scripts. For example, a few languages belonging to the Chinese and Japanese script share a common subset of characters. A word is more likely to be Japanese if both Chinese and Japanese characters occur in an image. But when no Japanese character occurs, the word has to be Chinese. We propose a sequence to sequence mapping of individual characters’ shape to their script, from which we can know what scripts the character of a given image could be. After that, the majority of the predicted probability distributions is calculated,and the dominant script is determined.

In this paper, we propose a novel scene text script identification method which uses spatial attention based sequence learning approach for script identification in scene text images. We use spatial attention blocks which extract rich text features while ignoring the disruptions present in the scene text. Inspired by the successful application of residual connections in image classification tasks \cite{he2016deep}, \cite{szegedy2017inception}, we employ hierarchical residual connections to build a deeper network that helps in learning representative image features. In training phase, instead of learning a single representation of the text and failing to capture the discriminating chars of each script, a segmentation free, sequence to sequence mapping is done using Connectionist Temporal Classification (CTC) to supervise the predictions. The key contributions of our scene text script identification method are: 

1. We propose a novel end-to-end trainable scene text script identification network which integrates spatial attention mechanism and residue convolutional blocks with the sequence learning capabilities of the LSTM to generate sequence to sequence mapping for discriminative learning of the script. The module has low latency and low memory footprint and thus is feasible for deployment in on-device text recognition systems. 

2. A deep feature extractor with residue convolutional blocks generates a rich image representation for the text.

3. The proposed sequence learning based script identification method does supervision using Connectionist Temporal Classification (CTC). To the extent of our knowledge, this is the very first time that CTC demonstrates strong capabilities for scene text script identification. 

\begin{figure}[b]
\begin{center}
  \includegraphics[width=\textwidth,height=\textheight,keepaspectratio]{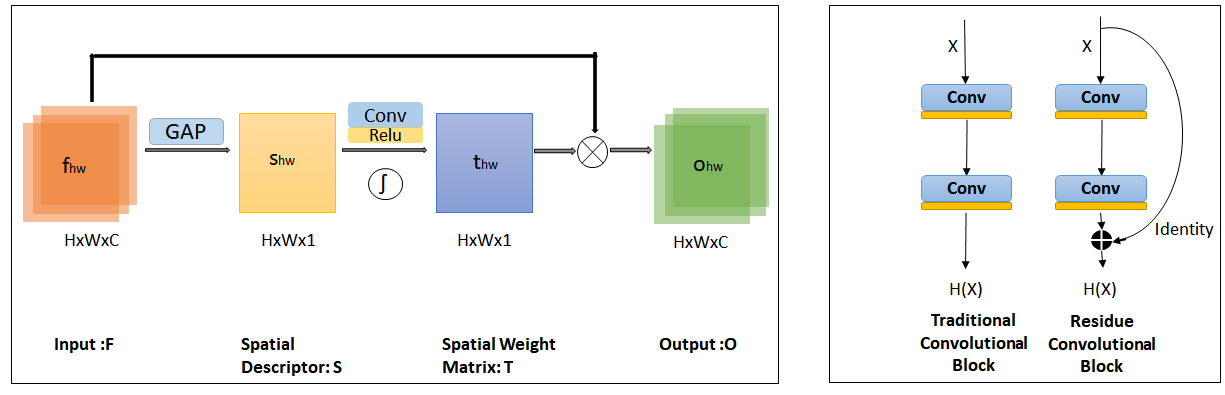}
\end{center}
 \caption{Illustration of the Spatial Attention Block (left) and the Residue Convolutional Block (right) which are the components of the feature extractor.}
\label{fig:arch}
\end{figure}

\section{Methodology}

In the following sections, we describe in detail the three key components of our network, namely, the spatial attention based feature extractor (\ref{sec:feature}), sequence modelling (\ref{sec:seq}) and prediction (\ref{sec:pred}). Figure \ref{fig:pipeline} illustrates the overall architecture of our network.

\subsection{Spatial Attention based Feature Extractor}
\label{sec:feature}
The feature extractor consists of a combination of two modules – Spatial Attention Block and the Residue Convolutional Block (see Figure \ref{fig:arch}). \\

\noindent
{\bf Spatial Attention Block}  Spatial attention mechanism in a convolutional neural network (CNN) ignores the information provided by the channels and treats all the channels in a feature map equally. In the CNN, the low level feature maps having fewer channels, are responsible for extracting the spatial characteristics such as edges and contours. Hence in our network, the spatial attention block is introduced for the low level feature maps to learn the interaction of spatial points and focus on the key informative areas while ignoring the irrelevant distortions. 

As can be seen from Figure  \ref{fig:arch}, to compute the spatial attention, firstly, we pass the feature map $F \in R^ {H\times W\times C}$ to the channel aggregation operation which generates a global distribution of spatial features, the spatial descriptor $S \in R^ {H\times W}$. Global average pooling $G_{avg}$ in the channel dimension (C) is used as the aggregation function. 
\begin{equation}
    s_{hw} \ = \ G_{avg}(f_{hw}) \ = \ \frac {1}{C}\sum _{i = l}^{C} f_{hw}(i)
\end{equation}
where $f_{hw} \in R^ {C}$ refers to the local feature at spatial position (h, w).
On the generated spatial descriptor $S$, weight learning is implemented using a convolutional layer 
to generate the spatial attention map $T \in R^{H\times W}$ which is computed as:
\begin{equation}
    T \ = \ \sigma (\delta (f_{1}p))
\end{equation}
where p is spatial position, $f_{1}$ represents a convolution operation with filter size $3\times 3$, $\delta$ refers to activation function ReLU \cite{nair2010rectified} and $\sigma$ is a sigmoid activation function used to generate spatial weight $t_{h, w} \in (0, 1)$ at position (h, w).  
T is finally applied to the feature map F. The feature values at different positions in F are multiplied by different weights in T to generate the output O of the spatial attention block as:
\begin{equation}
    O_{hw} \ = \ f_{hw}\cdot t_{hw}
\end{equation}

\noindent
{\bf Residue Convolutional Block}  The residue block (see Figure \ref{fig:arch} (b)) consists of two convolutional layers, each followed by ReLU activation and a skip-connection between the input of the first and the output of the second convolutional layer. If the input x is mapped to the true distribution H(x), then the difference can be stated as:
\begin{equation}
R(x) \ = \ H(x) - x  
\end{equation}

The layers in a traditional network learn the true output H(X) while the residue convolutional block layers learn the residue R(x), since they have an identity connection reference due to x. Other than the fact that it is easier to optimise the residual mapping than the original unreferenced mapping, an added advantage is that it can learn the identity function by simply setting the residual as zero. This enables to address the problem of vanishing gradients, giving us the ability to build deep feature extractor consisting of 7 convolutional layers.

\subsection{Sequence Modelling}
\label{sec:seq}
Previous works fail to take the spatial dependencies within text word into consideration, which may be discriminative for script identification as mentioned in section \ref{sec:intro}. 

Long Short-Term Memory (LSTM) \cite{hochreiter1997long} solves this problem by introducing three gates: input gate, forget gate, output gate. They are integrated together to capture spatial dependencies for a long time by keeping the gradient in the same order of magnitude during back-propagation. 
The hidden state $h_{t}$ depends not only on the current input $x_{t}$, but also previous time step hidden state $h_{t-1}$. 

In this work, we use one Bidirectional-LSTM (Bi-LSTM) layer with 256 hidden states which calculates $h_{t}$ using the memory block from both forward and backward directions. For faster inference, we use LSTM with recurrent projection layer \cite{sak2014long} to decrease the computational complexity. The projection layer projects the hidden state to a lower dimension of 96 at every time step. 

\subsection{Prediction}
\label{sec:pred}
We concatenate the features generated by the Bi-LSTM layer
and the per-frame predictions are fed into a fully connected layer, followed by a  softmax layer which outputs the normalized probabilities for each script class. 

We use Connectionist Temporal Classification (CTC) \cite{graves2006connectionist} to help the model associate the shape of the character to a script class. 

The formulation of the conditional probability is described as follows. Let $L$ represent the set of all script classes in our task and $L^{'} = L\cup \{blank\}$. The output from the sequential modelling block is transformed into per-frame probability distribution $y = (y_{1}, y_{2}..y_{T})$ where $T$ is the sequence length. Each $y_{t}$ is the probability distribution over the set $L'$. A sequence-to-sequence mapping function $\beta$ is defined on the sequence $ \pi \in L'^{T}$  which deletes repeated characters and the blank symbol. The conditional probability is calculated by summing the probabilities of all paths mapped onto the label sequence $l$ by $\beta$ : 

\begin{equation}
    p(l|y) \ = \ \sum _{\beta(\pi) = l} p(\pi |y)
\end{equation} 

As directly computing the above equation is computationally expensive, forward–backward algorithm \cite{graves2006connectionist} is used to compute it efficiently. While predicting the sequence of labels, we pick the label with the highest probability at each time step and then apply $\beta$ to the entire output path.  

During inference, we calculate the number of times each script is predicted for the given word crop. The script having the highest count is considered as the script for the word crop. 

\begin{table}[t]
\begin{center}
\begin{tabular}{|c|c|c|c|c|c|}
\hline
Method & MLT17 & CVSI 15 & MLe2e  & Parameters & Time
 \\
\hline\hline
Bhunia\cite{bhunia2019script}	& \bf{90.23}	& 97.75      & 96.70	   & 12	    & 85\\
ECN \cite{gomez2016boosting} 	& 86.46	        & 97.20	     & 94.40	   & 24	    & 13\\
Cheng\cite{cheng2019patch}	& 89.42	        & {\bf98.60}	     & -	   & 26.7	& {\bf2.5}\\
Mei\cite{mei2016scene}		& -             & 94.20	     & -	   & 5.2	& 92\\
Zdenek\cite{zdenek2017bag} 	& -	            & 97.11      & - 	   & -	    & 60\\
\hline
Ours 	& 89.59		    & 97.44      & {\bf97.33}   & {\bf1.1}    & 2.7\\

\hline
\end{tabular}
\end{center}
\caption{Script Identification accuracy on benchmark datasets. The best performing result for each column is set in bold font. The number of parameters are in millions (M) and time is denoted in milliseconds per image (ms).}
\label{tab:comp}
\end{table}
\section{Experiments}
\subsection{Dataset}

We evaluate our scene text script identification methods on following benchmark datasets:

{\bf ICDAR 2019} \cite{nayef2019icdar2019} (IC19) consists of 8 script classes - Latin, Korean, Chinese, Japanese, Arabic, Hindi, Bangla and Symbols. The test set consists of approximately 102462 cropped word images and is highly biased towards Latin. 

{\bf RRC-MLT 2017} \cite{nayef2017icdar2017} (MLT17) consists of 97,619 test cropped images. This dataset holds 7 scripts: Arabic, Latin, Chinese, Korean, Japanese, Bangla, Symbols. There exist some multioriented, curved and perspectively distorted texts which make it challenging.

{\bf CVSI 2015} \cite{sharma2015icdar2015} (CVSI 15) dataset contains scene text images of ten different scripts: Latin, Hindi, Bengali, Oriya, Gujrati, Punjabi, Kannada, Tamil, Telegu, and Arabic. Each script has at least 1,000 text images collected from different sources (i.e. news, sports etc.)

{\bf MLe2e} \cite{gomez2016fine} is a dataset for the evaluation of scene text end-to-end reading systems.
The dataset contains 643 word images for testing covering four different scripts: Latin, Chinese, Kannada and Hangul many of which are severely corrupted by noise and blur and may have very low resolutions.

\subsection{Implementation Details}

For feature extraction, the input text crops are scaled to fixed height of 24 while maintaining the aspect ratio. The residue convolutional blocks are stacked with the spatial attention block being introduced after the 1st and 2nd block as shown in Figure \ref{fig:pipeline}. The filter size, stride and padding size of all the convolutional layers is 3, 1 and 1 respectively and each convolutional layers is followed by a max-pooling layer. The output filter size for the convolutional layers are 32, 64, 96, 128, 164, 196, 256 respectively and each layer is followed by the ReLU activation function. Batch normalization \cite{ioffe2015batch} is adopted after the 3rd and the 5th convolutional layer, speeding up the training significantly. The input layer is followed by a 3 × 3-stride max-pool layer. This is followed by three residual blocks, with each unit consisting of two 3×3 convolution operations. The first residual block further downsample the feature maps with a 2 × 2-stride max-pool layer, while the last two residual blocks each employ a 2 × 1-stride max-pool layer to reduce the sizes of feature maps only in the height dimension. The two LSTMs of a BiLSTM layer in the encoder each have 256 hidden units for capturing dependencies in each direction of the feature sequence. 

We initialize the network by first training on a synthetic dataset generated using the method followed by \cite{gupta2016synthetic}. The training set consists of 1.5 million word crops with equal distribution for all the scripts, which are randomly sampled to form minibatches of size 96. Further, the network is fine-tuned on the benchmark datasets and evaluated. We adopt some data augmentation techniques such as varying contrast, adding noise and perspective transformations to make the dataset robust. The Adam optimizer is used with the initial learning rate being set to 0.001 and is automatically adjusted by the optimizer during training. The training of the network on Nvidia Geforce GTX 1080ti GPU, 16GB RAM takes about 36 hours for 10 epochs. The network has approximately 1.1 million parameters and takes 2.7 ms per image for identification.

\begin{table}[t]
\begin{center}
\begin{tabular}{|c|c|c|}
\hline
Module & Accuracy & Params \\
\hline\hline
CNN+LSTM               & 84.79 & 0.65 \\
ResCNN+LSTM            & 84.66 & 1.0\\
Atten+CNN+LSTM         & 86.75 & 1.03\\
Atten+ResCNN+LSTM      & 88.60 & 1.1\\
Atten+CNN+LSTM+CTC     & 87.85 & 1.03\\
Atten+ResCNN+LSTM+CTC  & 90.17 & 1.1\\
\hline
\end{tabular} 
\begin{tabular}{|c|c|}
\hline
Position of SA blocks & Accuracy 
 \\
\hline\hline
After conv and 1st RB	&  83.81	\\
After conv and 2nd RB	&  84.59	\\
After 1st and 2nd RB	&  85.63	\\
After 2nd and 3rd RB	&  84.81	\\

\hline
\end{tabular}
\end{center}
\caption{(a) Comparison with different variations of the proposed network on MLT17 validation set. Parameters are mentioned in millions. (b) Script Identification accuracy on MLT17 with respect to the position of spatial attention blocks, where RB stands for Residual Block and SA stands for Spatial Attention Block}
\label{tab:spaattn}
\end{table}

\subsection{Effectiveness of Spatial Attention and Residue Convolutional Blocks for Script Identification}

To validate the effectiveness of the proposed spatial attention based residue convolutional network for scene text script identification, in Table \ref{tab:spaattn} (a), we compare the performance with that of a baseline CNN network without the application of the spatial attention mechanism or residue convolutional blocks but with the same sequence modelling backbone. Specifically, in the ’CNN+LSTM’ model we skip the spatial attention block and residual connections, feeding the input text image into the 'CNN+LSTM' backbone and using cross entropy classification as the loss function. The model ‘ResCNN+LSTM’ omits the spatial attention block and makes use of the residue convolutional blocks, enhancing the quality of image feature representation. On the other hand, the model 'Atten+CNN+LSTM' skips using the residual connections and employs the spatial attention block, using cross entropy as the optimization function while 'Atten+CNN+LSTM+CTC' makes use of the CTC loss function.The model ’Atten+ResCNN+LSTM’ combines the spatial attention blocks with the residue convolutional blocks, and the model ’Atten+Res+CNN+LSTM+CTC’ instead uses CTC loss function to optimise the prediction. 

{\bf Contribution of spatial attention blocks and residue convolutional blocks} Compared to the baseline model, the introduction of the proposed spatial attention module and the residue convolutional module effectively increases the accuracy of the script identification model.  Combining both modules helps further enhance the performance on all the datasets. 'Atten+ResCNN+LSTM' attains more accurate residual enhancement representations as compared to 'Atten+CNN+LSTM', thus capturing low level features, which help improve the identification accuracy. The overall enhanced identification accuracy of the proposed spatial attention based sequence learning approach for script identification relative to the baseline, demonstrates the effectiveness of spatial attention for scene text script identification. 


{\bf Arrangement of spatial attention blocks} To ensure low latency and computational complexity we limit to using two spatial attention blocks. Given an input image, the two attention blocks compute attention, focusing on exploiting image features while ignoring distortions. Considering this, the two blocks can be placed at either the low level or at the higher level in the network. In this experiment we compare four different ways of arranging the two spatial attention blocks: After conv and the 1st residual block, after conv and 2nd residual block, after the 1st and 2nd residual block and finally, after the 2nd and 3rd residual block. Since the spatial attention works locally, the order may affect the overall performance. Table \ref{tab:spaattn} (b) summarises the experimental results on different combinations. From the results we can find that spatial attention blocks after the 1st residual block and the 2nd residual block is slightly better than the other arrangements. 

{\bf Effect of CTC loss} To verify the superiority of the proposed sequence to sequence mapping method for script identification, which performs per character script classification, we replace the proposed CTC module in our script identification model ’Atten+ Res+ CNN+ LSTM+ CTC’, with the Cross entropy classification technique, the model ’Atten+ Res+ CNN+ LSTM’, which generates a single representation for the input image and the spatial attention based residual network is trained. The experimental results show that CTC module yields 1.57{\%} higher script identification accuracy on the benchmark dataset compared to the cross entropy module whose output features may be ignoring the discriminative features between similarly structured scripts.

\begin{table}[t]
\begin{center}
\begin{tabular}{|c|c|}
\hline
Method & Accuracy (\%)
\\
\hline\hline
Tencent - DPPR team      & 94.03\\
SOT: CNN based classifier     & 91.66\\
GSPA\_HUST    & 91.02\\
SCUT-DLVC-Lab   & 90.97\\
TH-ML & 88.85 \\
Multiscale\_HUST     & 88.64\\
Conv\_Attention      & 88.41\\
ELE-MLT based method     & 82.86\\
\hline
Ours    & 89.82\\
\hline
\end{tabular} 
\begin{tabular}{|c|c|}
\hline
Script & Accuracy 
\\
\hline\hline
Latin      & 98.30\\
Korean     & 94.28\\
Chinese    & 90.28\\
Japanese   & 84.15\\
Arabic     & 92.61\\
Hindi      & 86.28\\
Bangla     & 87.28\\
Symbols    & 85.44\\
\hline
Average    & 89.82\\
\hline
\end{tabular}
\end{center}
\caption{(a) Comparison of our results on MLT 19 with other methods. (b) Script wise results of our method on MLT19 dataset.}
\label{tab:mlt19}
\end{table}
\subsection{Comparison with State-of-the-Art Script Identification Methods}
Our method achieves the highest script identification accuracy on one of the three benchmark datasets and a highly competitive accuracy with the last two datasets. Note that our network is significantly lighter and faster as compared to other state-of-the-art methods which qualifies it for real-time on-device pre-ocr applications. 

For the scene text images in Mle2e, a great improvement has been made. Since in this dataset, many of the text regions are severely distorted and few of them have very low resolutions, the performance of our network shows the ability and the need for a spatial attention mechanism in handling disruptions occurring in natural images. As can been seen from the Table \ref{tab:comp}, our method also performs well on the MLT17 dataset.Table \ref{tab:cvsi} compares our method with others on CVSI 2015 dataset. The result demonstrates that our method performs well on both document text images and complex background images. 

Table \ref{tab:mlt19} mentions script wise performance accuracy of our method ICDAR19 RRC (IC19). Table \ref{tab:mlt19} also compares the performance of our method with other models described in \cite{nayef2019icdar2019}. The baseline methods have based their methods on famous deep nets for text recognition such as ResNet, VGG16, Seq2Seq with CTC etc. and use the recognised characters to identify the scripts while adopting improvements such as multiscale techniques, voting strategy for combining results from multiple nets, training statistics of the scripts etc. Even though the proposed network is slightly less accurate the benchmarks, it is superior in terms of latency and model size and thus, in contrast, is feasible for on-device deployment.

Given the relatively standard feature extraction backbone employed in our method, the results of the experiment demonstrates the significant effect of the proposed spatial attention mechanism coupled with the sequence learning capabilities of LSTM on improving the accuracy of the script identification model. Particularly, compared to the existing script identification methods, Section \ref{sec:intro}, which either use complex self-attention mechanisms or heavy feature extractors, our method with a simple convolution based attention block and a small CNN backbone achieves overall enhanced accuracy, with the additional benefit of being computationally lighter and faster.


\begin{table}
\begin{center}
\begin{tabular}{|c|c|c|c|c|c|c|c|}

\hline
Script  & C-DAC & HUST & CVC-2 & CIJK & Cheng & {\bf Ours}
 \\
\hline\hline
Eng   & 68.33 & 93.55 & 88.86 & 65.69 & 94.20 & 98.54\\
Hin   & 71.47 & 96.31 & 96.01 & 61.66 &  96.50 & 97.24\\
Ben   & 91.61 & 95.81 & 92.58 & 68.71 &  95.60 & 97.74\\
Ori   & 88.04 & 98.47 & 98.16 & 79.14 & 98.30 & 98.16\\
Guj   & 88.99 & 97.55 & 98.17 & 73.39 & 98.70 & 96.94\\
Pun   & 90.51 & 97.15 & 96.52 & 92.09 & 99.10 & 97.47\\
Kan   & 68.47 & 92.68 & 97.13 & 71.66 & 98.60 & 97.45\\
Tam   & 91.90 & 97.82 & 99.69 & 82.55 & 99.20 & 96.88\\
Tel   & 91.33 & 97.83 & 93.80 & 57.89 &  97.70 & 96.60\\
Ara   &97.69 & 100.00 & 99.67 & 89.44 & 99.60 & 97.36\\
\hline
Average  & 84.66&  96.69 & 96.00 & 74.06 & 97.75 & 97.44\\

\hline
\end{tabular}
\end{center}
\caption{Script wise performance comparison with various methods on CVSI 2015 dataset.}
\label{tab:cvsi}
\end{table}
\section{Conclusion}
In this paper, we present a novel lightweight real-time on-device spatial attention based sequence learning approach for scene text script identification. The spatial attention mechanism helps removing spatial disruptions in natural images for improving script identification accuracy in complex scenarios. Our network employs residue convolutional blocks to generate rich feature representations of the input image thus capturing discriminative features of a script. Sequence learning using the Bidirectional LSTM layer, optimised using the Connectionist Temporal Classification enables the model to associate the characters shape to its script, thereby enhancing script identification accuracy.  While achieving competitive accuracy with state-of-the-art methods, our network is superior, as it achieves 4 times reduction in model size and has low latency, making it suitable for on-device inference. 

%
%
%
\bibliographystyle{splncs04}
\bibliography{references}
%




\end{document}